\documentclass[runningheads]{llncs}
\usepackage{graphicx}
\usepackage{geometry}
\usepackage{url}
\usepackage{epstopdf}
\usepackage{color}

\begin{document}
\title{A 1d convolutional network for leaf and time series classification}
\author{Dongyang Kuang}
	%orcidID{0000-1111-2222-3333} 
	%	\and
	%	Tanya Schmah\inst{2}
%\orcidID{1111-2222-3333-4444}
%%index{Kuang, Dongyang}
%%index{Schmah, Tanya}
\authorrunning{D. Kuang et. al}
%% First names are abbreviated in the running head.
%% If there are more than two authors, 'et al.' is used.
%%
\institute{University of Ottawa, Ottawa, Canada.\\
	\email{dykuangii@gmail.com} \\
	%	University of Ottawa,Ottawa, Canada. \\
	%	\email{tschmah@uottawa.ca}
}
\maketitle
%\linenumbers
\begin{abstract}
	In this paper, a 1d convolutional neural network is designed for classification tasks of plant leaves. This network based classifier is analyzed in two directions. In the forward direction, the proposed network can be used in two ways: a classifier and an automatic feature extractor. As a classifier, it takes the simple centroid contour distance curve as the single feature and achieves comparable performance with state-of-art methods that usually require multiple extracted features. As a feature extractor, it produces nearly linear separable features, hence can be used together with other classifiers such as support vector machines to provide better performance. The proposed network adopts simple 1d input and is generally applicable for other tasks such as classifying one dimensional time series in an end-to-end fashion without changes. Experiments on some benchmark datasets show this architecture can provide classification accuracies that are comparable or higher than some existing methods. In the backward direction, methods like gradient-weighted class activation mapping and maximum activation map of neurons in the classification layer with respect to inputs are performed to help investigate and further validate that hidden signatures helping trigger the trained classifier's specific decisions can be human interpretable. Code for the paper is available at 
	\url{https://github.com/dykuang/Leaf_Project.}
\end{abstract}

\section{Introduction}
There are vast amount of plant species existing on earth. According to previous research such as \cite{species1,species}, flowering plants alone have over millions of different species. This amount of species and the requirement of fast identification in modern applications bring a huge challenge upon traditional taxonomy methods. While the problem of large in-species variations and small cross-species variations in this context are already challenging enough, expert-level knowledge and vast human labor are also required for making correct classifications of plant species. Being faced with these challenges, the fast growing machine learning and deep learning algorithms seem to provide a more suitable solution from a different direction, especially for non-experts.

From a descriptive point of view, plant identification are traditionally based on observations of its organs, such as flowers, leaves, seeds, etc. A large portion of species information is contained in plants' leaves. Leaves also exist for a relative long time during plants' life cycle. Traditionally, features from leaves can be roughly divided into three categories: shape, color and texture. Shape descriptors (especially the contour) usually are more robust compared to the other two. For a single leaf, color descriptors may vary depending on lighting conditions, image format, etc. Texture descriptors can also vary if the leaf are damaged...  Another advantage of a shape descriptor is that features like centroid center contour curve (CCDC) can be viewed as time series \cite{shapelet}, hence techniques in time series classification such as dynamic time warping (DTW) \cite{dtw} can be applied. On the other hand, techniques that are suitable for leaf classification with this kind of shape descriptor can be easily modified to general time series classification tasks, which will result in a broader field of applications. 

Despite differences of features, traditional classifiers in applications usually includes: support vector machines (SVM), k nearest neighbors (kNN), random forest ... Artificial neural networks, especially convolutional neural networks (CNN) \cite{krizhevsky2012imagenet} are not commonly seen in the field, though they have proven to be very effective tools for applications in the field of computer vision and pattern recognition. In this paper, discussions are focused on features that are based on leaf shapes. The author argues that simple shape feature actually contains more discriminating power than people usually think if an effective classifier such convolutional neural networks are used. In the forward direction, the proposed network takes simple 1 dimensional easily extracted feature and allows an nearly end-to-end classification, so that it can be conveniently deployed in actual applications. The use of simple single feature as CCDC helps improve the general applicability of the constructed network for different tasks than it is originally designed for. The trained network can also work as an universal feature extractor allowing more in-depth process down the stream.  In the backward direction, the trained network can help post-analysis such as the question about what features in the input help trigger the specific decision of the classifier.

%{\color{red} More details about each section. More about general applicability.}
The rest of the paper is organized as below: Section \ref{sec:RW} gives some related work using shape features for classification. Section \ref{sec:cls} presents the design of a 1d convolutional network as a classifier for the simple shape feature that is based on contours. Section \ref{sec:exp} tests the performance of this network directly as a classifier and also as a feature extractor/processor combined with other classical classifiers such as support vector machine or $k$ nearest neighbors on some benchmark data sets from different sources. To support the general applicability of proposed method, it is also applied to the task of time series classification which the classifier was not originally designed for. The section also examines the trained network via visualizing the learned features and pattern that help trigger the trained networks's decisions in the classification layer.

\section{Related Work}
\label{sec:RW}
Effort for developing classification tools can generally be divided into two parts: feature engineering that extracts more discriminative, meaningful and interpretable features and classifier design that are more efficient, accurate and generally applicable to similar tasks.

On the side of shape features, they can be extracted based on botanical characteristics \cite{BotChar,BotChar1} or via other feature engineering efforts. The botanical features may include geometrical measurements such as aspect ratio, rectangularity, convex area, ratio, convex perimeter ratio, sphericity, circularity, eccentricity, form factor, etc. Other non botanical handcrafted features with multi-scale or hierarchical properties are also commonly seen in the literature. \cite{trig} discussed some other features applied on leave shapes and introduced two new multi-scale triangle representations. There are also a lot of other work done with more in-depth design aiming for general shapes than just leaves. \cite{innerdistance} defines inner distance of shape contours to build shape descriptors.  \cite{centrist} develops the visual descriptor called CENTRIST (CENsus TRansform hISTogram) for scene recognitions, it gets good performance when applied to leave images. Authors of \cite{shapelet} uses the transformation form shape contours to 1 dimensional time series and present the method of shapelet for shape recognition. \cite{HMDS} describes a hierarchical representation for two dimensional objects that captures shape information at multiple levels of resolution for matching deformable shapes. \cite{wang2015march} focuses on mobile retrieval task with multiscale-arch-height description as feature. \cite{zhao2015plant} takes a pattern counting approach free of preprocessing of input data. \cite{yang2016multiscale} develops a multi-scale triangular centroid distance feature for the recognition. Features coming from different methods can also be ensembled, these bagged features can usually help provide better performance as discussed in \cite{adinugroho2018leaves,UCI100} or \cite{kaur2019plant}. These works spend major efforts in hand crafting features that will benefit particular classification tasks. There are also methods based on classical artificial neural networks such as \cite{adinugroho2018leaves,kho2017automated} or modern convolutional neural networks such as \cite{he2016leaf,jeon2017plant}. These work either use neural network as a replacement of traditional classifiers for adopting ensembled features or are based directly on 2 dimensional leaf images that allows an end-to-end classification but implicitly takes in a lot of other features than the single shape information. 

This paper intend to use a single type feature of centroid center contour curve (CCDC)for classification. This contour curve is a feature that is derived from a relatively easy concept and can be efficiently/conveniently extracted from leaf images. The simple input feature also helps perform an easier analysis of what the trained network is looking at via methods such as activation maximization \cite{erhan2009visualizing} than ensembled or hierarchical features.
Some early work \cite{ccd1,ccd} used this feature as the single feature or in addition to other features. To the author's best knowledge, it was rarely used (at least not as a single feature) in recent years because people doubt that it alone by itself may not have enough discriminative power. The author argues that if a classifier is designed properly, it can reveal more hidden information out of CCDC and provide comparable or better performance when compared to some state-of-art methods mentioned above. 

To obtain CCDC representation, one first apply a filter such as a canny filter \cite{canny} on the image to obtain the leave contour. For point $(x, y)$ on this contour, its polar coordinates $(\rho, \theta)$ is then computed:
\begin{eqnarray}
\rho &=& \sqrt{(x-x_0)^2 + (y-y_0)^2}\\
\theta &=& \arctan{\frac{y-y_0}{x-x_0}}.
\end{eqnarray}
$(x_0 , y_0)$ is the image center and can be computed from image moments \cite{moment}. Values of $\rho$ then can be sampled on a uniform grid of $\theta$ by interpolation. CCDC is obviously translation invariant. It can also be rotation and scale invariant after proper normalization. 
\begin{figure}[tbh]
	\centering
	\includegraphics[width=0.45\textwidth]{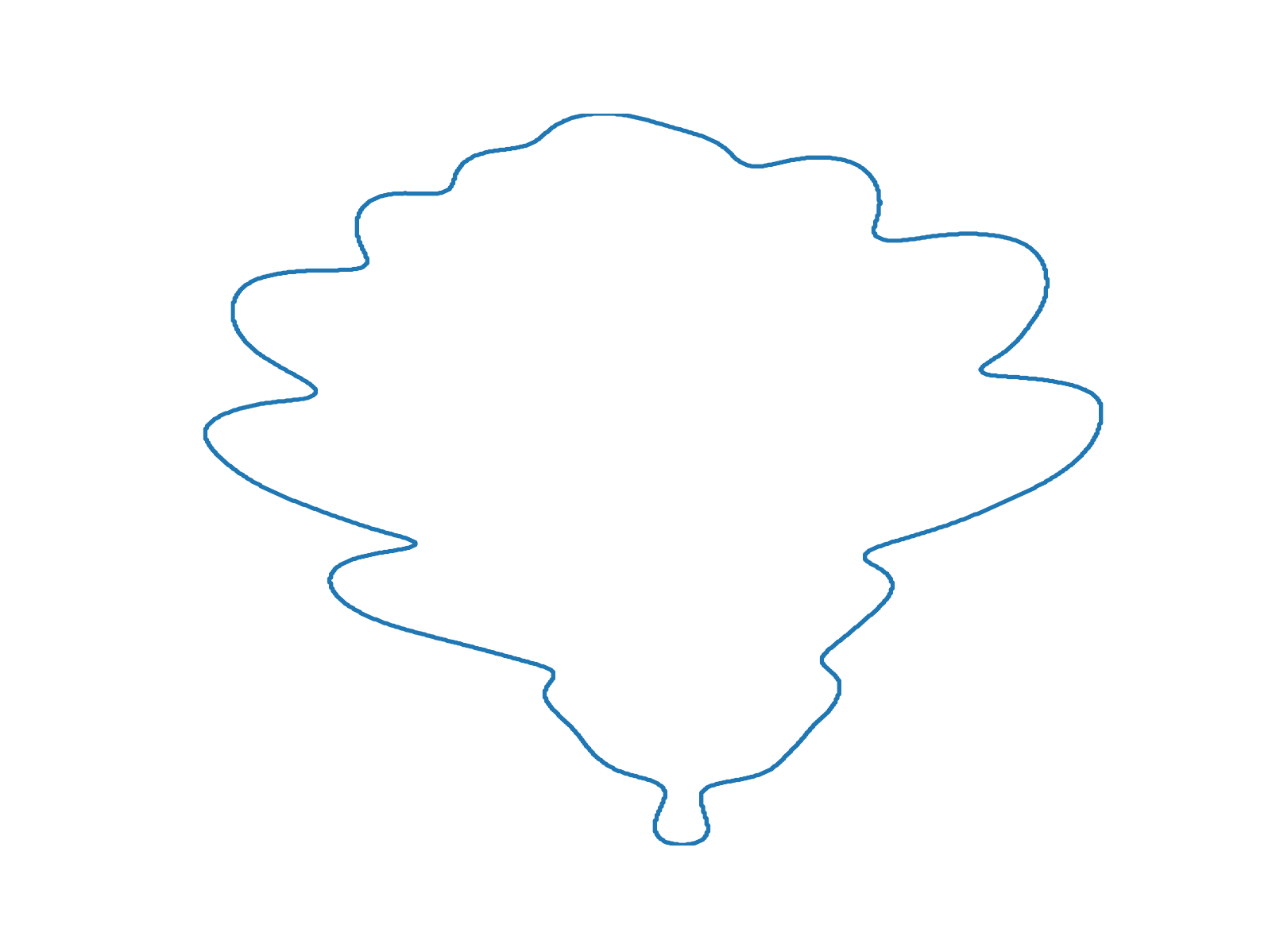}
	\includegraphics[width=0.45\textwidth]{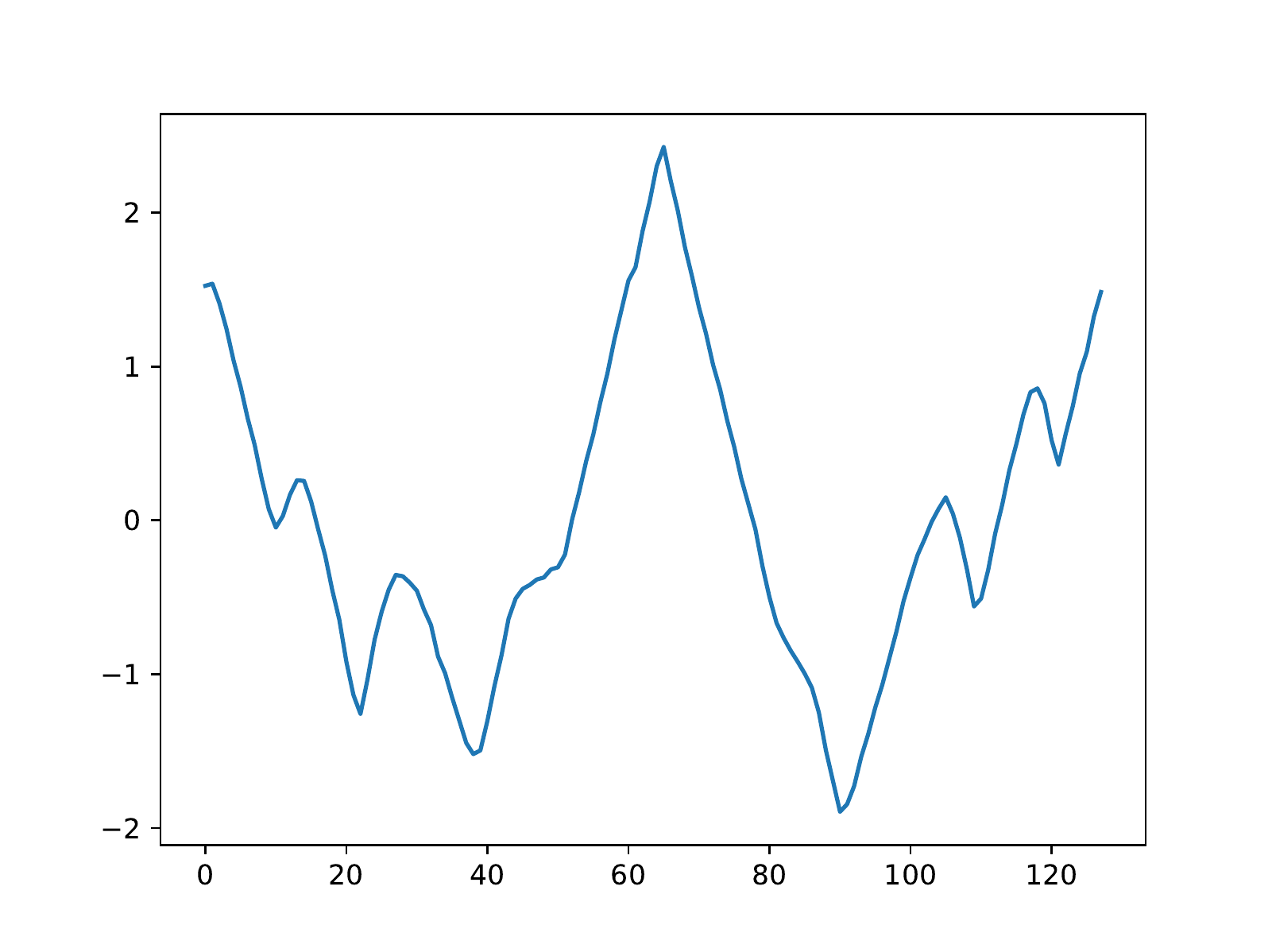}
	\caption{An example of CCDC. Above: Outline of one Quercus leaf; Below: the converted CCDC.}
\end{figure}

Compared with methods mentioned above which tackles the difficulty in classification by designing complicated hand crafted deep features, convolutional neural networks (CNN) \cite{krizhevsky2012imagenet} can take simple features as input and automatically abstracts useful features through its early convolutional blocks for later classification tasks \cite{zeiler2014visualizing}. In this way, the difficulty is transferred into heavy computation where modern hardware now can provide sufficient support. It is more straightforward if we apply a CNN  directly on leave images combining feature extraction task and classification task together, but this will make a model of unnecessary large size with a lot of parameters and they usually require a lot of data and time to be trained well with more risk of overfitting the data at hand. The key idea of this paper is to take the advantage of convolutional architecture, but apply it on the extracted single 1d CCDC feature to reduce the computational cost. The CCDC as a simpler and more general representation also help transfer the built network for other tasks such as classifying time series. The success on the same network on different problem domains will also help confirm the effectiveness and general applicability of the proposed classifier.

\section{Classifier Design}
\label{sec:cls}
\begin{figure}[tbh]
	\centering
	\includegraphics[scale=0.5]{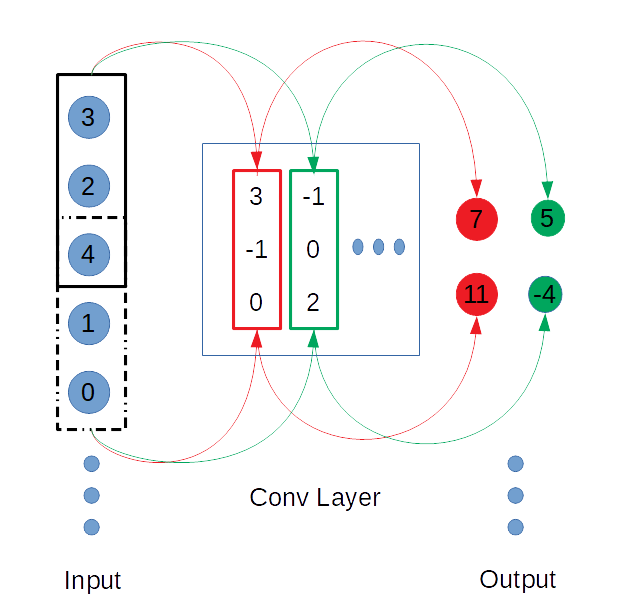}
	\caption{Mechanism of a 1d convolutional layer. Kernel weights, i.e. digits in the red/green rectangles are learned during training.}
	\label{fig:CONV1D}
\end{figure}
%{\color{red} In this section, add more details of terminology such as BN,
%activation, dropout. Aspects of the architecture optimization}
In order to make proper classification, it is important that the classifier can learn features at different scales together and combine them into classification. Though this can be done by designing complicated hand-crafted features as mentioned in section \ref{sec:RW}, applying convolutional kernels with different sizes and strides and let it learns by itself from presented samples also serves as one good option for this purpose. For a typical 1d convolutional mechanism, information flows to the next layer first by a convolutional operation and then processed by an activation function: $Y = f(W*X+B)$, where 
$*$ denotes the discrete convolution operation between the incoming signal $X$ and a kernel $W$. A convolutional layer contains several different kernels or filters, computes the convolution between the input and each kernel and then stack their result as its output. Unlike filters such as Gaussian filter or Canny filter that have predetermined convolutional kernels serving for specific purpose, kernel weights used in a neural network are post-determined, they are gradually learned during the training task, i.e. the result of optimization algorithm aiming to minimize an object function. Taking the simplest case for example, if some $Y_0$ is a desired value, the weight $W$ and $B$ is determined from the following minimization problem:
\begin{equation}\label{eq:min}
{MIN}_{\{W, B\}} \quad L(f(W*X + B), Y_0)
\end{equation}
where $L$ is some loss function measuring the difference between the current prediction $f(W*X + B)$ and the desired value $Y_0$. 
Figure \ref{fig:CONV1D} gives an illustration of the forward pass,
the convolutional layer contains several kernels of length 3. During convolution, a sliding window of the same size will slide through the input with certain stride. During each stay of the window, it computes the inner product between the examined portion of input and the kernel itself.  For example, when using kernel $(3,-1,0)$ with stride 2 and no bias, the first output is $3\times 3 + 2 \times (-1) + 4 \times 0 = 7$ and the second output is $4\times 3 + 1 \times (-1) + 0 \times 0 = 11$. In the following backward pass, the kernel $(3,-1,0)$ then will be updated according to problem (\ref{eq:min}). A full 1d convolutional network will contain several convolutional layers of this kind and possibly other layers.

\begin{figure*}[tbh]
	\centering
	\includegraphics[width=\textwidth]{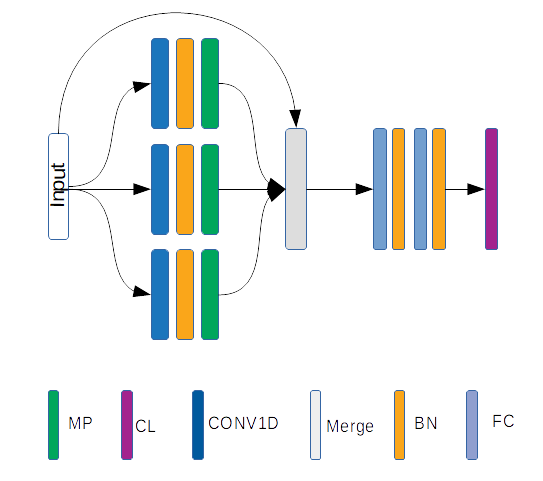}
	\caption{The architecture of the neural network classifier. The right most layer is a classifier layer (CL). It can be a linear classifier, a (kernel) SVM classifier, a knn classifier or other classifiers. The merge layer is simply concatenation of features. Batch normalization (BN) \cite{ioffe2015batch} layer can be asserted after the output of convolutional or full connected layer (FC) to help better training. The three convolutional layers are with \textit{different} sizes and strides.}
	\label{fig:nn}
\end{figure*}

Based on this thought, a basic architecture used for classification is designed as in Figure \ref{fig:nn}. It looks like a naive module from Google's inception network \cite{inception} but is built for 1 dimensional input. The input is first processed by convolutional blocks of different configurations in parallel which aims to capture and highlight to features of different scales separately. Their outputs are then concatenated together with original input of full resolution before being fed into latter layers for classification. ReLU (Rectified Linear Unit) activation function : 
\begin{equation}
\Phi(x) = \textit{max}(0, x)
\end{equation} 
in convolutional layers to help accelerate the training. Parametric ReLU (PReLU) \cite{prelu} activations:
\begin{equation}
\phi(x)=\left\{
\begin{array}{ll}
x & \quad x \geq 0 \\
\alpha x & \quad x < 0
\end{array}
\right.
\end{equation} 
are used for fully connected layers. The parameter $\alpha$ is learned during training. This kind of activation allows small gradient when the neuron is not active, hence helping avoid the ``dying ReLU" problem.
Batch normalization (BN) layers \cite{ioffe2015batch} are also utilized in the architecture design. BN can be understood as the transformation over each batch during training by $\mathrm{BN}(\gamma^{\ell},\beta^{\ell}): x^{\ell} \to y^{\ell}$. It normalizes each dimension $\ell$ of  input vector $x = (x_1,\ldots,x_n)$, via 
\begin{equation}
x_{\mathrm{norm}}^{\ell} = \frac{x^{\ell}-{E}[x^{\ell}]}{\sqrt{\mathrm{Var}[x^{\ell}]}}.
\end{equation} 
The value $x_{\mathrm{norm}}^{\ell}$ is used for the internal layer, and the layer \[y^{\ell} = \gamma^{\ell} x_{\mathrm{norm}}^{\ell} +\beta^{\ell},\] is passed to following network layers, where $\gamma^{\ell}$ and $\beta^{\ell}$ are scaling and shifting parameters learned. Research such as \cite{shimodaira2000improving} showed that 
batch normalization can help reduce the internal covariate shift in the network parameters.

In the last dense layer, softmax activation function: 
\begin{equation} 
\phi (x)_i = \frac{e^{x_i}}{\sum_{j=1}^c e^{x_j}}
\end{equation} 
is applied over $c$ classes for classification purpose. The output of this layer can be naturally interpreted as a probability distribution.  The whole network is then optimized using a categorical cross entropy loss function: 
\begin{equation}
{E}_{\mathrm{loss}} = \sum_{i=1}^{c}T_i\log (\phi(x)_i)
\end{equation}

The architecture design also involves several techniques for preventing the issue of possible overfitting. First, Gaussian noise layers are placed before each of the convolutional layers as a way of data augmentation. Dropout layers are placed before the last classification layer as a way of regularization to randomly drop out learned features, it help the network avoid learning features that are only subject to particular batch hence improve the over-fit issues.

In the following experiment section, this network is used in two ways. The first approach is to use it as a classifier allowing informations flow from CCDC feature to species label directly. The other way is to use it as an automatic feature extractor/processor in a ``pretrain-retrain" style. During the training phase, the network is first pretrained to certain extent with earlystopping or a checkpoint at best validating performance. In the testing phase, the model weights are frozen, the top layer is then taken off and its input as pretrained features are fed to a nonlinear classifier such as a SVM or a kNN classifier for final classification. It is like a transfer learning design, but the difference is in transfer learning, the model is not trained on the same dataset. The idea is from heuristic that a nonlinear classifier may performance better than the original linear classification performed by the top layer. Experiments done in the next sections shows this (referred as 1dConvNet+SVM) usually will help contribute a little more accuracy to the classification. 

\section{Experiments}
\label{sec:exp}
In order to show that the built network is effective and generally applicable for classification tasks, we used several publicly available datasets from different sources for benchmarks. We applied 10-fold cross validation in Section \ref{Swedish Leaf} with the Swedish leaf data set \cite{Swedish_leaf_data} for a robust evaluation. An 16-fold cross validation was used in Section \ref{100 Leaf} with UCI's 100 leaf dataset \cite{100leaf} for comparing with other reported methods in a consistent way. In addition, we also took one step further by trying the \textit{same} network on some time series classification tasks from UEA \& UCR time series classification repository \cite{UCR_ts} in Section \ref{timeseries}. The \textit{same} explicit split train and test set are used for the evaluation as the best methods reported in the website. The proposed network gives comparable or better performance than state-of-art in all these benchmarks.

\subsection{Swedish Leaf}
\label{Swedish Leaf}
Swedish leaf data set \cite{Swedish_leaf_data} contains leaves that are from 15 species. Within each species, 75 samples are provided. It is an challenging classification task due to its high inter-species similarity \cite{trig}. 
\begin{figure}[tbh]
	%	\centering	
	\includegraphics[width = \textwidth, trim = 0.75in 0in 0in 0in, clip]{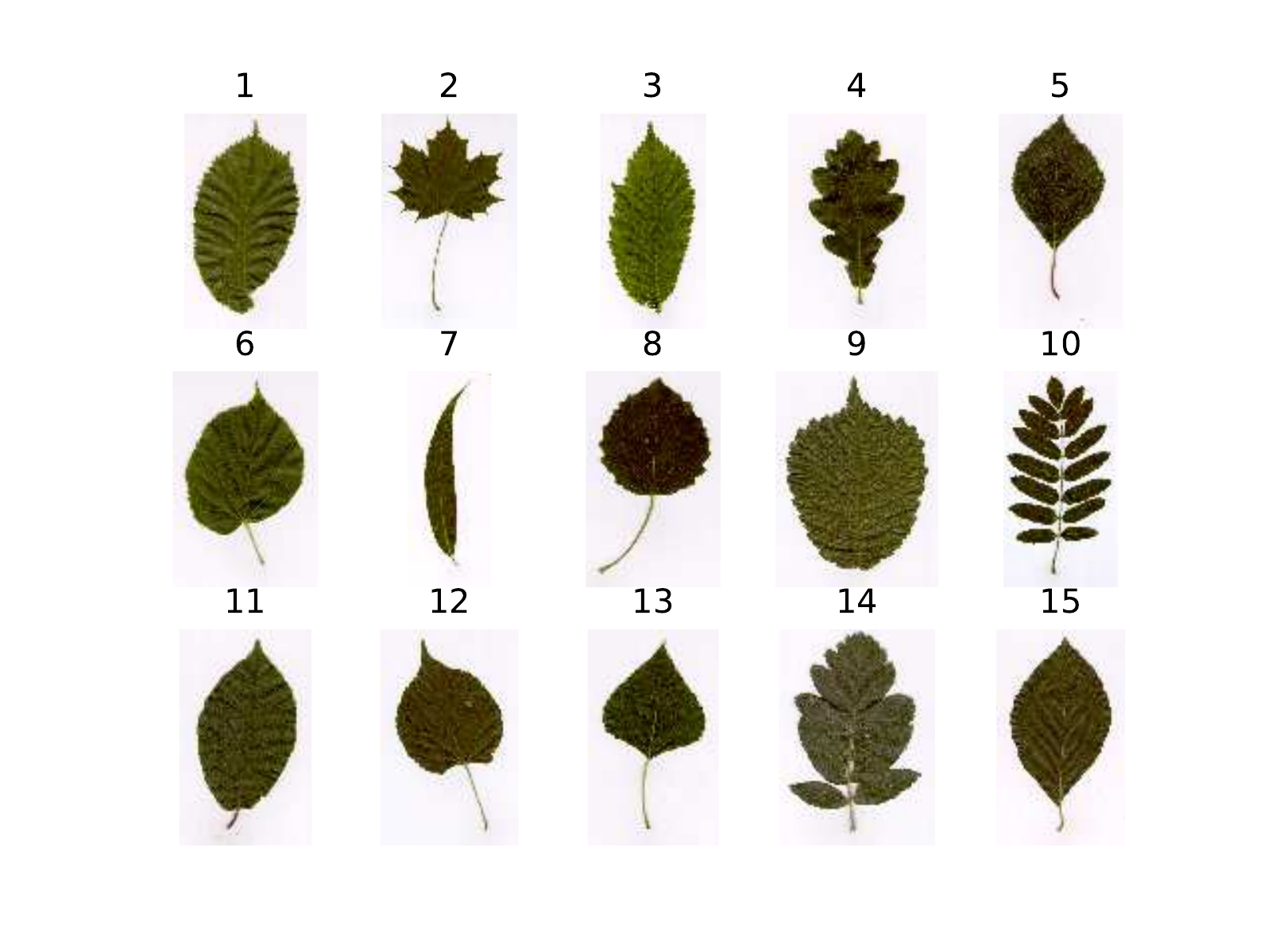}
	\caption{The first sample of each species in the Swedish leaf dataset. 1. Ulmus capinifolia, 2. Acer, 3. Salix aurita, 4. Quercus, 5. Alnus incana, 6. Betula pubescens, 7. Salix alba 'Sericea', 8. Populus tremula, 9. Ulmus glabra, 10. Sorbus aucuparia, 11. Salix sinerea, 12, Populus, 13. Tilia, 14, Sorbus intermedia, 15. Fagus silvatica}
	\label{fig:leaf}	
\end{figure}

Table \ref{tab:leaf_acc} lists some existing methods that used leaf contours for classification. All listed methods in the table used leaf contours in a non-trivial way that involves more in-depth feature extraction than CCDC.
\begin{table}[tbh]
	\centering
	\caption{Performance of different existing methods involving features from leaf contours.}
	\label{tab:leaf_acc}
	\begin{tabular}{|l|c|}
		\hline
		Method & Accuracy \\
		\hline
		S\"{o}derkvist \cite{swedish_leaf_origin} & 82.40\%\\ Spatial PACT \cite{centrist} & 90.61\% \\
		SC + DP \cite{innerdistance}                &88.12\%  \\ Shape-Tree \cite{HMDS}     & 96.28\% \\
		IDSC + DP \cite{innerdistance}               &94.13\%  \\ TSLA \cite{trig}            & 96.53\%\\ 
		MTCD \cite{yang2016multiscale}            & 96.31\%\\ 
		MSVM \cite{kaur2019plant}            & 93.26\%\\
		I-IDSC + NN \cite{zhao2015plant}  &  97.07\%\\
		MARCH + NN \cite{wang2015march} & 97.33\%\\
		\hline
		
	\end{tabular}
	
\end{table}

While \cite{HMDS,innerdistance,trig,swedish_leaf_origin,centrist} uses 25 samples randomly selected from each species as the training set and the rest as test. The author decided to use a 10-fold cross validation to evaluate the proposed model in a more robust way. The other reason for this is the convolutional architecture may not be trained sufficiently with 25 samples per species as the training set. The mean performance and the corresponding standard deviation is summarized in Table \ref{tab:10cv}. The actual parameters used are:
Convolutional layers \{conv1d(16, 8, 4)\footnote{16 kernels with window size 8 and stride 4.},  conv1d(24, 12, 6), conv1d(32, 16, 8)\}, Maxpooling layers (MP) are with window size 2 and stride 2, two fully connected layers are of unit 512 and 128, respectively.% Relu activations \cite{relu} are used in convolutional layers and PRelu \cite{prelu} activations are used for fully connected layers. 
To prevent overfitting, Gaussian noise (mean: 0, std: 0.01) layers are placed before each convolutional layer and a dropout layer \cite{JMLR:v15:srivastava14a} of intensity 0.5 is inserted before the classification layer. The whole model is trained using stochastic gradient descent algorithm with batch size 32, learning rate 0.005 and $10^{-6}$ as the decay rate. 25 principal components from pretrained features are used if the top classification layer is a SVM. For other details, please check the actual code at \cite{code}.

\begin{table}[tbh] 
	\centering
	\caption{Performance of the 10-fold cross validation using the 1d ConvNet directly as the classifier and also as an auto feature extractor with the last layer replaced by other classifiers such as NN or SVM.
	}
	\label{tab:10cv}
	\begin{tabular}{|l|c|c|c|c|}
		\hline
		Method & Mean Acc. & STD & Best & Worst\\
		\hline
		1d CNN & \textbf{96.11\%} & 1.54\% &  \textbf{98.23\%} &  92.92\%  \\
		1d CNN + 3NN & \textbf{94.69\%} & 1.58\% & \textbf{96.46\%}& 91.15\%\\
		1d CNN + SVM & \textbf{97.08\%} & 1.48\% & \textbf{99.12\%} & 94.69\%\\
		\hline
	\end{tabular}
\end{table}
The proposed network provides comparable accuracy with top methods listed in Table \ref{tab:leaf_acc}. With a SVM on pretrained features from the network, it is able to provide a better accuracy. A 3NN classifier on the same pretrained features does not give better performance in this experiment. It should be noted here that the SVM and NN used in the table and following experiments are not specially tuned, they serve as an example that the network when used as feature extractor can also be effective when combined with other simple classifiers. 

The UEA \& UCR Time Series Classification Repository \cite{UCR_ts} provides an explicit split of training/test set of this dataset and a list of performances from different time series classification methods, which allows a more direct comparison with the proposed 1d convolutional network.  Table \ref{tab:leaf_cote} lists the best performance reported on the website and results obtained by the proposed 1d ConvNet. The result is obtained by averaging the test accuracy among 5 independent runs with different random states. 20\% of the training samples are used as validation for stopping the training process\footnote{Unless specified otherwise, accuracies recorded in the rest experiments of this paper is obtained with the same way.}.
\begin{table}[tbh]
	\centering
	\caption{Performance comparison on the explicit training/test split from the UEA \& UCR Time Series Classification Repository. }
	\label{tab:leaf_cote}
	\begin{tabular}{|l|c|}
		\hline
		Method &  Accuracy \\
		\hline
		COTE \cite{cote} & 96.67\%\\
		1dConvNet & \textbf{96.10\%}\\
		1dConvNet+3NN & \textbf{96.16\%}\\
		1dConvNet+SVM & \textbf{97.47\%}\\
		\hline
	\end{tabular}
\end{table}

\begin{figure}[tbh]
	\centering
	\includegraphics[width=\textwidth, trim=1.25in 0 0 0 ,clip]{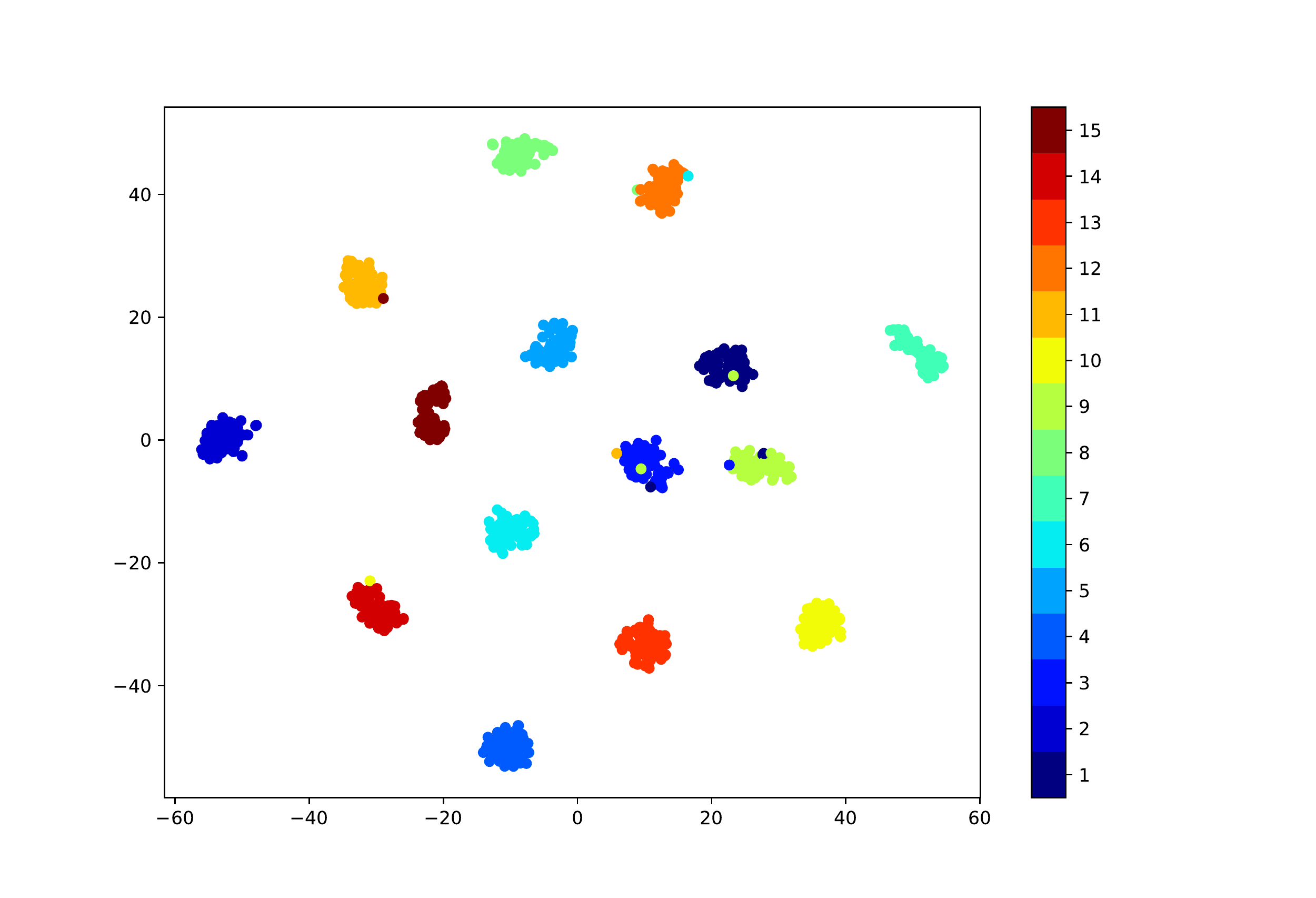}
	\caption{TSNE embedding of the whole dataset using the inputs from the classification layer. The 15 classes are almost linear separable.}
	\label{fig:tsne}
\end{figure}

As seen in both comparisons, with top layers replaced by a SVM, the accuracy can be further improved. The reason may be the fact that if the network is already trained properly, information that flows into the top layer is almost linearly separable, hence a nonlinear classifier built on top will help increase the accuracy by correcting some mistakes made by a linear classifier. In order to further demonstrate that the trained network actually learned features that are almost linearly separable before the last dense classification layer, we used TSNE embedding \cite{tsne} to project the high-dimensional feature into a 2 dimensional visualization. Figure \ref{fig:tsne} shows these projected features, the 15 classes are almost separable. 

\begin{figure}[tbh]
	\centering
	\includegraphics[width=\textwidth]{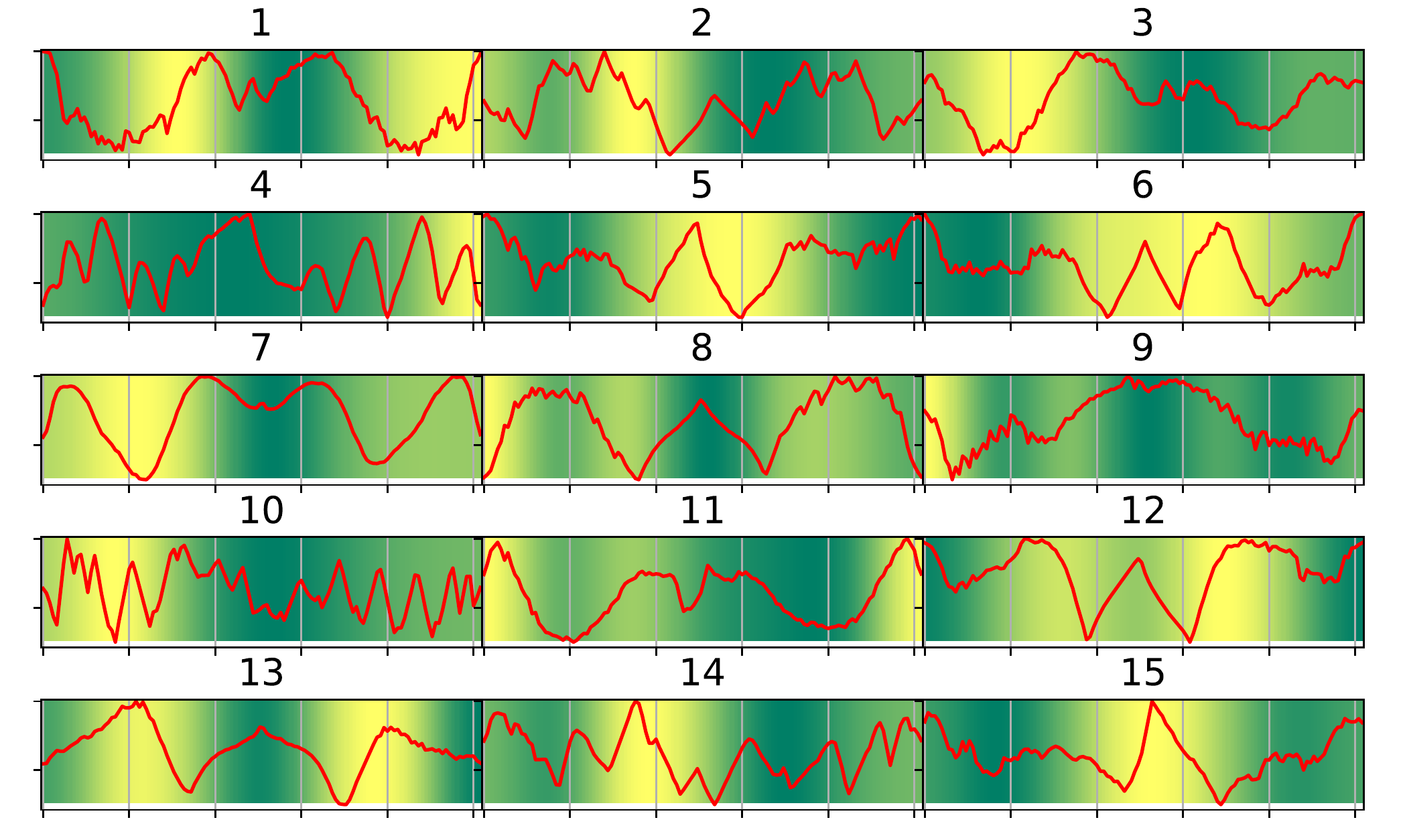}	
	\caption{Normalized CCDC curves overlaid with attention map when the trained network is making predictions on each sample per class. The two ends of the subplots correspond to the part near the leaf tip, the center corresponds to the part of curve near the leaf stem.}\label{fig:cam}
\end{figure}

In the forward direction, the classifier is able to perform good classifications. A natural question then arises that can the trained classifier help backward analysis? Given a class label, what kind of information in input CCDC help trigger the trained neural network's specific label prediction? It is possible that the trained network just picked up some non-related sample bias instead of general existing patterns that are human interpretable for classification. 
To answer this question, we utilized two methods upon the trained network as part of the backward analysis. First, we examined the class activation map \cite{cam} via gradient-weighted class activation mapping(Grad-CAM) algorithm \cite{gradCAM} as a way of visualizing attention over input when the trained network performs predictions. The idea is to use the normalized gradient information of the output or other specified concept with respect to inputs for producing a coarse localization map that highlights parts in the input. This result can be intuitively viewed as an relative attention/contribution map of the input towards the output through the neural network classifier. We show the result in Figure \ref{fig:cam} by overlaying the attention map and the normalized 1D input CCDC curve. It can be observed from the figure that sharp corners or turns receives more attention in most cases when the network tries to make a prediction. Combining with sample images from Figure \ref{fig:leaf}, we can make some preliminary observations. 

There are cases like labels 12, 13 where attention maps are separated as two main symmetric locations given the very symmetric leaf shapes. For label 1, two major attentions are not symmetric, this may because the leaf shape is not quite symmetric. Attention map for a sample from label 8 is also symmetric, but the network seems to put slightly more attention on the tip area. Labels like 9 and 11 have a relative narrow attention map on leaf tip. Labels like 3, 7, 10 have a broader attention on one side close to the left tip area. Label 6 has a attention map well over most all of the leaf. Label 5 and 15 shares attention mostly on the center.
Leaf labels 4, 10 and 14 are compound leaves, they share the similar behavior that the one major attention is covering windows where changes of amplitude are largest. These similarities of attention maps provide a different angle of visualizing the small cross-species variations and is one of the reasons behind misclassification of the trained network.

\begin{figure}
	\centering
	\includegraphics[width=\textwidth]{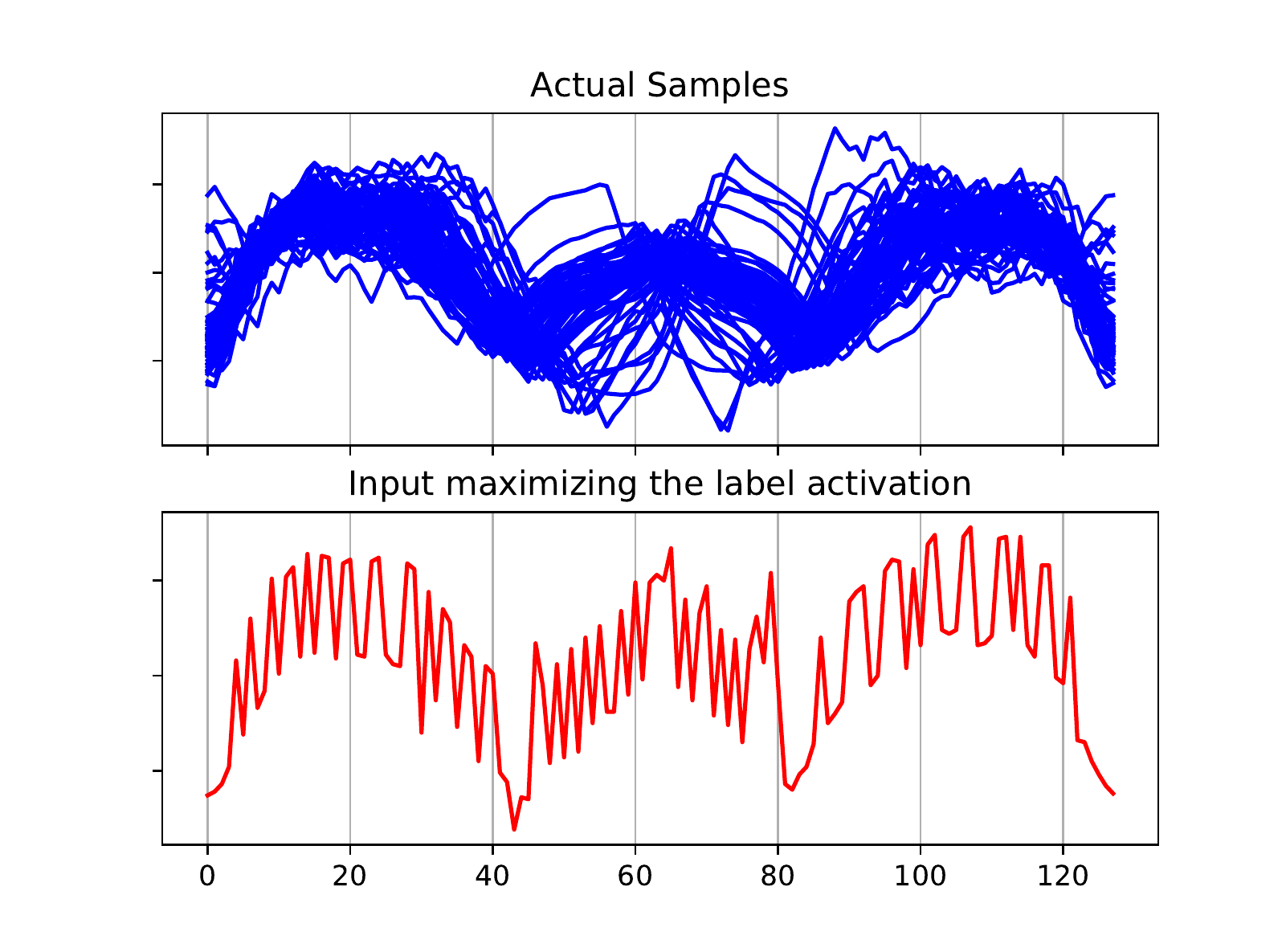}	
	\caption{Sample CCDC from Populus tremula compared with the learned template from neural network. Top: an overlay of all CCDC samples from training. Bottom: The input maximizing the activation corresponding to Populus tremula label with the trained neural network.}\label{fig:AM}
\end{figure}

Second, we took the idea of activation maximization in \cite{erhan2009visualizing} to examine the trained network. The idea is simple, parameters within the trained neural network will be frozen at first. An optimization will then be conducted to search for input CCDCs that maximize the neuron's activity corresponding to the given label. The calculation is done effectively by a Keras visualization toolkit \cite{raghakotkerasvis}. We show one case for the label: Populus Tremula in Figure \ref{fig:AM}. In this figure, we compare the overlay of all CCDC samples in the training set from the given label and the ``template" learned from the activation maximization algorithm. The template is certainly not a faithful recovery of what a \textit{particular} Populus Tremula leaf would look like, it is instead a certain kind of \textit{abstract summary or exaggeration} of class features learned from the \textit{whole} presented samples during training. The band-like trend formed by small and frequent spikes can also be viewed as the result from the large in-species variance. For the case shown in Figure \ref{fig:AM}, it can be observed at first glance that there are at least two places the trained network is looking at when classifying the input as Populus Tremula: the global trend and the local ``random" varying spikes.  

\begin{figure}
	\centering
	\includegraphics[width=\textwidth]{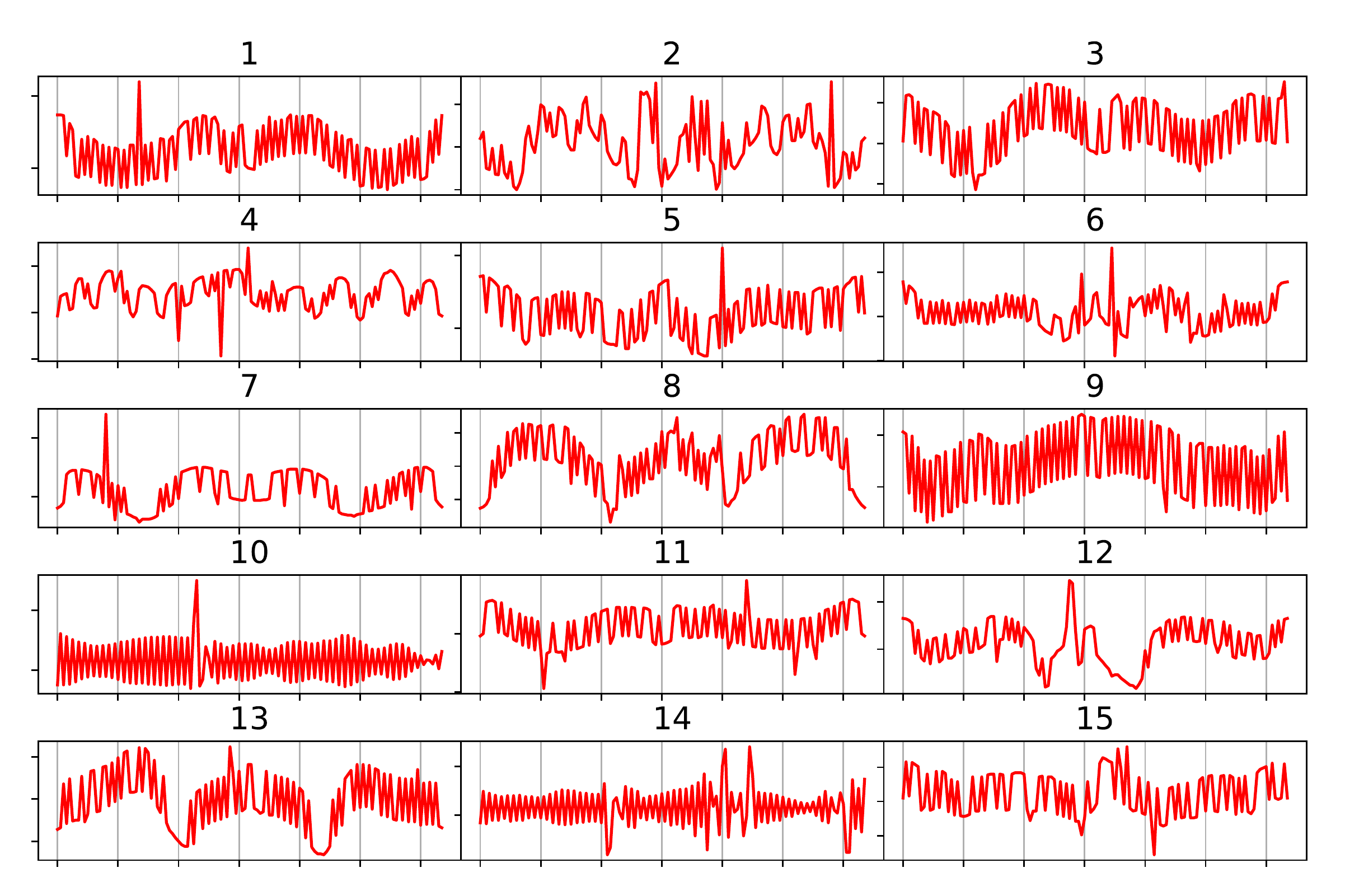}	
	\caption{Input maximizing the activation corresponding to each label from the dataset with the trained neural network.}\label{fig:AM_all}
\end{figure}

This example together with Figure \ref{fig:cam} provide supports that the trained network indeed is looking at meaningful features for its assigned classification task. Specifically, features that the network put its eyes on can also be human interpretable. This observation is confirmed by visualizing inputs that maximizes other labels separately with the same trained network as seen in Figure \ref{fig:AM_all}. 

\subsection{UCI's 100 leaf}
\label{100 Leaf}
UCI's 100 leaf dataset \cite{100leaf} was first used in \cite{UCI100} in support of authors' methods about probabilistic
integration of shape, texture and margin
features for better classifications. It has 100 different species with 16 samples per species\footnote{One sample's texture feature from the first species is missing, so actually data from the other 99 species is used in this experiment.}. It is a somewhat more challenging task for neural network approaches since it contains more classes and less samples per sample. As for the feature vector, a 64 element vector is given per sample of leaf. These vectors are taken as a contiguous descriptors (for shape) or histograms (for texture and margin). An mean accuracy of 62.13\% (with proportional density estimator \cite{mallah2013plant}) and 61.88\% (with Weighted PROPortional density estimator \cite{mallah2013plant}) was reported by only using the shape feature(CCDC) from a 16-fold validation (10\% of training data are hold as validation). The mean accuracy raised up to 96.81\% and 96.69\% if both three types of features are combined. We will use the same features here and performance two kinds of comparison: one with only the CCDC feature and the other with all the three features.
Following the evaluation of 16-fold cross validation used by the original paper, the performance of using the 1d ConvNet is summarized in Table \ref{tab:uci}. For results by combing the 3 features, the author simply concatenates them together to form a 192 dimensional feature vector per sample instead of dealing with them separately. 

\begin{table}[tbh]
	\centering
	\caption{Comparison of performance on UCI's 100 leaf dataset.}
	\label{tab:uci}
	\begin{tabular}{|l|c|c|}
		\hline
		Method & CCDC & All 3 features\\
		\hline
		PROP\cite{mallah2013plant} & 62.13\% & 96.81\%\\
		WPROP\cite{mallah2013plant} & 61.88\% & 96.69\%\\
		1dConvNet&\textbf{73.99\% $\pm$ 3.72\%}&\textbf{99.05\% $\pm$ 0.67\%}\\
		1dConvNet+3NN&\textbf{73.86\% $\pm$ 3.66\%}&\textbf{98.73\% $\pm$ 1.41\%}\\
		1dConvNet+SVM&\textbf{77.34\% $\pm$ 3.55\%}&\textbf{99.43\% $\pm$ 0.62\%}\\
		\hline
	\end{tabular}
\end{table}
Again, the proposed network works better on both kinds of input features. The 3-NN with pretrained features from the network performed worse than the original network. Part of the reason may be because kNN classifier is more sensitive to changes in data and 3 may not be a good choice for $k$ in this dataset which has 99 different classes. 

\subsection{Time series classification}
\label{timeseries}
%{\color{red} Rewrite the intro to avoid going into time series classification but as an example of general applicability.}

The single CCDC feature can be viewed as an particular example of one dimensional time series. In order to demonstrate the proposed architecture is generally applicable in end-to-end 1d time series classification tasks, the author selects four different data sets from UEA \& UCR Time Series Classification Repository \cite{UCR_ts}: ChlorineConcentration, InsectWingbeatSound, DistalPhalanXTW and ElectricDevices\footnote{These datasets come with explicit split of train and test set. Details of these data can be found at the website \cite{UCR_ts}.} for test. These datasets come from different backgrounds with different data sizes, different number of class labels and different feature vectors. A good classification strategy usually requires some prior knowledge for effective feature extraction and selection. With the help of convolutional architecture, the proposed network is able to help reduce such prior knowledge from human. This kind of prior knowledge is ``learned" by the network during training.
The current best performance reported on the website and performance achieved by this 1d convolutional net are compared in Table \ref{tab:comp}.  For all the four datasets, the network's architecture and hyperparameters are \textit{the same} as previous experiments with no extra hyperparameter tuning\footnote{For the DistalPhalanXTW dataset, the author took 10\% of them as validation.}. As summarized in Table \ref{tab:comp}, the proposed network outperforms all the four reported best methods in terms of mean accuracy. It is noted here that the author has no intention to compete in the field of time series classification, but instead taking this different task as an example to show the general applicability of proposed method. Performance will be better if domain specific prior knowledge of the input are added or cooperated in the method design.

\begin{table}[tbh]
	\centering
	\small
	\setlength\tabcolsep{1pt}
	\caption{Performance achieved by the proposed 1d convolutional netwrok compared to reported best performance on \cite{UCR_ts}. Acronym used: Support Vector Machine with Quadratic kernel(SVM-Q), Random Forrest(RF), ShapeletTransform(ST) \cite{st}}
	\label{tab:comp}
	\begin{tabular}{|l|c|l|c|}
		\hline\noalign{\smallskip}
		Dataset & Classes & Best Method Reported& 1dConvNet+SVM\\
		\hline\noalign{\smallskip}
		ChlorineConcentration & 3 & 84.57\%  \quad \textit{SVM-Q} &  \textbf{99.77\%}\\
		InsectWingbeatSound & 11 &  63.89\%  \quad \textit{RF}&  \textbf{76.61\%}\\
		ElectricDevices &  7 &89.54\%  \quad \textit{ST}  & \textbf{94.34\%}\\
		DistalPhalanXTW &  6 &69.32\%  \quad \textit{RF}  & \textbf{71.22\%}\\
		\hline
	\end{tabular}
\end{table}

\section{Conclusion}
This paper presents a simple one dimensional convolutional network architecture for plant leaf classification tasks. The architecture allows nearly end-to-end classifications on single easily extracted CCDC feature instead of complicated, hand-crafted and domain specific features. The proposed network can also work a universal feature extractor that allows further in-depth processes such as stacking another downstream classifier to help better performance. Beyond the task of leaf classification, the proposed architecture is generally applicable for classifying one dimensional time series \textit{without} changes as a baseline approach. Experiments of this classifier on several benchmark datasets show comparable or better performance than other existing state-of-art methods, visualizations on learned features and visualization tools like gradient weighted class activation map and  activation maximization are also conducted for verifying the trained network has indeed learned useful patterns for classification. There are still many questions remained to be answered from this work. For example, the mapping between machine interpretable features and human interpretable features. The answer of these hard questions are waited to be discovered by future research.

\section*{Acknowledgement}
The author thanks Prof. Tanya Schmah and Dr. Alessandro Selvitella from University of Ottawa for their kind help in providing many useful suggestions.

\bibliographystyle{unsrt}
\bibliography{leaf}

\end{document}